\documentclass[letterpaper, 10 pt, conference]{ieeeconf}  

\IEEEoverridecommandlockouts                              

\usepackage{cite}
\usepackage{amsmath,amssymb,amsfonts}
\usepackage{bm}
\usepackage[caption=false, font=footnotesize]{subfig}
\usepackage{algorithm}
\usepackage{algorithmic}
\usepackage{graphicx}
\usepackage{textcomp}
\usepackage{xcolor}
\usepackage{tikz}
\let\labelindent\relax
\usepackage{enumitem}
\usepackage{hyperref}
\usepackage{todonotes}
\usepackage[symbol,hang,flushmargin]{footmisc}
\usepackage[normalem]{ulem}
\usepackage[nolist]{acronym}
\usepackage{listings}
\usepackage{float}
\usepackage{comment}
\usepackage{subfig}

\acrodef{spi}[SPI]{Shadow Program Inversion}
\acrodef{dof}[DoF]{degree of freedom}
\acrodefplural{dof}[DoF]{degrees of freedom}
\acrodef{tcp}[TCP]{tool center point}
\acrodef{dmp}[DMP]{Dynamic Movement Primitive}
\acrodef{dcg}[DCG]{differentiable computation graph}
\acrodef{xai}[XAI]{explainable artificial intelligence}
\acrodef{dgpmp2}[dGPMP2]{Differentiable Gaussian Process Motion Planning}
\acrodef{dgpmp2-nd}[dGPMP2-ND]{Differentiable Gaussian Process Motion Planning for N-DoF Manipulators}
\acrodef{gpmp}[GPMP]{Gaussian Process Motion Planning}
\acrodef{dp}[$\partial$P]{differentiable programming}
\acrodef{sdf}[SDF]{signed distance field}
\acrodef{gpu}[GPU]{graphics processing unit}
\acrodef{spi-dp}[SPI-DP]{Shadow Program Inversion with Differentiable Planning}
\acrodef{gp}[GP]{Gaussian process}

\definecolor{lbcolor}{rgb}{0.95,0.95,0.95}  
\title{\LARGE \bf
Shadow Program Inversion with Differentiable Planning: A Framework for Unified Robot Program Parameter and Trajectory Optimization
}

\author{Benjamin Alt$^{1,2,*,\dagger}$, Claudius Kienle$^{1,*}$, Darko Katic$^{1}$, Rainer Jäkel$^{1}$ and Michael Beetz$^{2}$%
\thanks{This  work  was  supported  by  the  German  Federal  Ministry  of  Education and Research (grant 01MJ22003B), the DFG CRC EASE (CRC \#1320) and the EU project euROBIN (grant 101070596).}%
\thanks{$^*$Equal contribution}
\thanks{$^\dagger$Corresponding author: {\tt benjamin.alt@uni-bremen.de}}
\thanks{$^{1}$ArtiMinds Robotics, Karlsruhe, Germany}%
\thanks{$^{2}$Institute for Artificial Intelligence, University of Bremen, Germany}%
}

\begin{document}

\maketitle
\thispagestyle{empty}
\pagestyle{empty}

\begin{abstract}
This paper presents \acf{spi-dp}, a novel first-order optimizer capable of optimizing robot programs with respect to both high-level task objectives and motion-level constraints. To that end, we introduce \acf{dgpmp2-nd}, a differentiable collision-free motion planner for serial N-DoF kinematics, and integrate it into an iterative, gradient-based optimization approach for generic, parameterized robot program representations. \ac{spi-dp} allows first-order optimization of planned trajectories and program parameters with respect to objectives such as cycle time or smoothness subject to e.g. collision constraints, while enabling humans to understand, modify or even certify the optimized programs. We provide a comprehensive evaluation on two practical household and industrial applications.
\end{abstract}

\section{Introduction}
\label{sec:introduction}

Skill-based robot programming has eased the use of robots for solving real-world applications. However, the cost of automation of complex tasks is often driven by the manual tweaking of robot trajectories and program parameters, such as approach poses, waypoints or force thresholds. This labor-intensive process requires skilled programmers and time-consuming trial and error. ``Programming by optimization'' \cite{hoos_programming_2012} addresses this issue by allowing a human programmer to specify a rough program skeleton, which is then completed by an optimization algorithm. This approach is particularly useful in tactile applications like force-controlled insertion or handling of deformable objects. However, applying general-purpose optimization methods to robot programs is challenging, as the success of one robot skill may depend on the parameterization of preceding skills: An optimizer must \textit{jointly} optimize the parameters of complete skill sequences or hierarchically composed subprograms. Moreover, robots must not only achieve task-level goals such as cycle time requirements, but also respect motion-level constraints such as collision-freeness or proximity to a human demonstration. Existing approaches focus exclusively on either trajectory optimization \cite{ratliff_chomp_2009,Kala11,osa_multimodal_2020,dastider_retro_2024} or parameter optimization \cite{racca_interactive_2020,alt_robot_2021,berkenkamp_bayesian_2023,kumar_practice_2024} and typically optimize individual skills, rather than jointly optimizing complete robot programs.

\begin{figure}
	\centering
	\includegraphics[width=\linewidth]{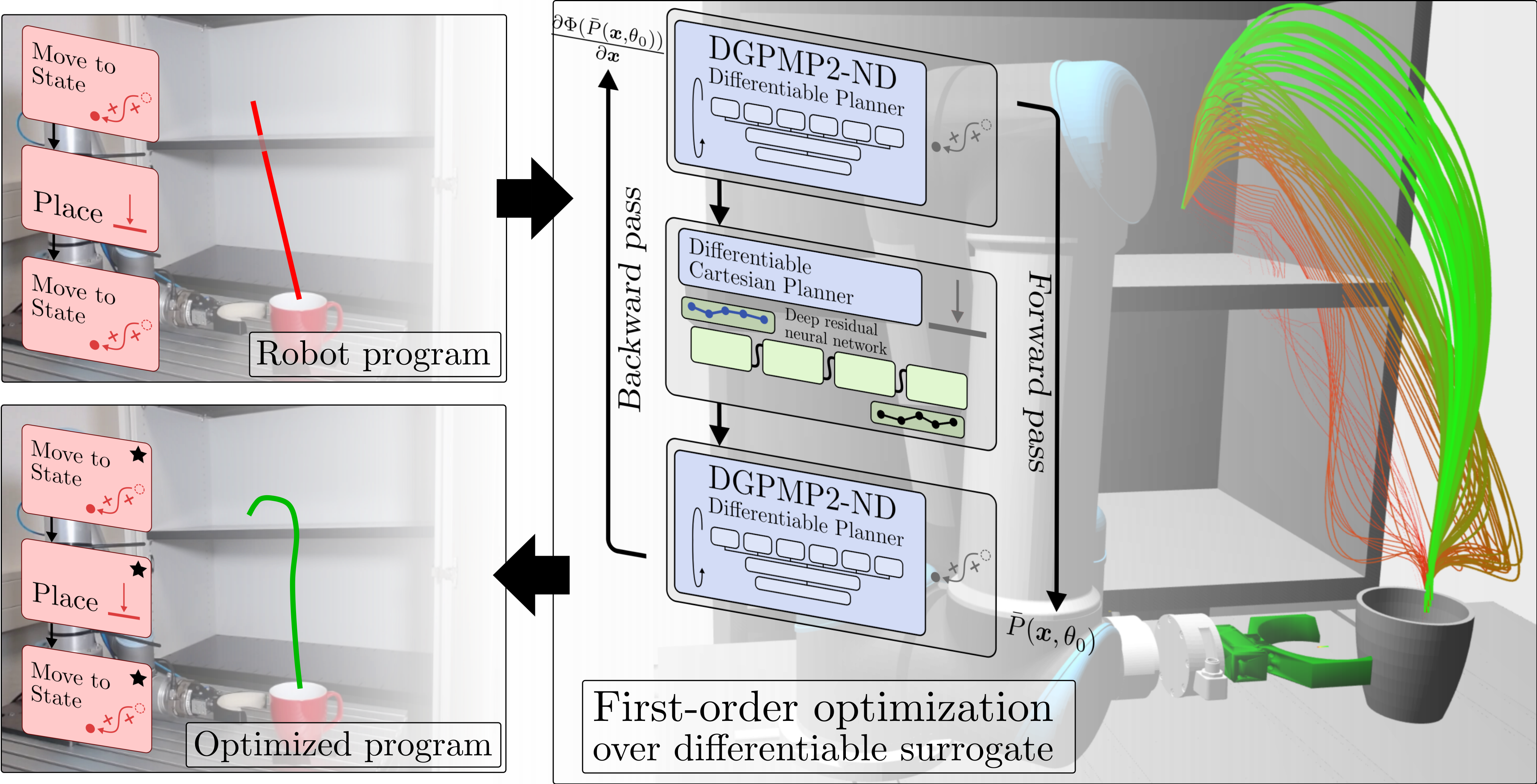}
	\caption{\acf{spi-dp} enables the optimization of robot programs (left, red) by first-order iterative optimization over a differentiable surrogate (right, gray). A differentiable collision-free motion planner (\acs{dgpmp2-nd}) ensures that the resulting motion trajectories are optimal with respect to task objectives and motion-level constraints.}
	\label{fig:overview_frontpage}
\end{figure}

In this paper, we propose \acf{spi-dp}, a robot program optimizer which combines both trajectory and parameter optimization with respect to task- and motion-level constraints. We make the following contributions:

\begin{enumerate}
	\item We present \textbf{\acf{dgpmp2-nd}}, a \textbf{differentiable motion planner} for serial N-\ac{dof} manipulators, capable of propagating gradients through the collision-free planning procedure.\footnote{The source code for \ac{dgpmp2-nd} is available at \url{https://github.com/benjaminalt/dgpmp2-nd}.}
	\item We introduce \textbf{\ac{spi-dp}}, an approach for \textbf{unifying program parameter and trajectory optimization} capable of optimizing robot programs with respect to a wide range of objective functions, including task-specific metrics and goals demonstrated by humans.
	\item We provide a \textbf{real-world evaluation} of the proposed framework on household pick-and-place as well as industrial peg-in-hole applications.
\end{enumerate}

To our knowledge, \ac{spi-dp} is the first approach to combine parameter and trajectory optimization for robot programs in one unified framework.

\section{Related Work}
\label{sec:related_work}

\subsubsection{Robot program parameter optimization}

In the context of ``programming by optimization'' \cite{hoos_programming_2012}, a wide array of optimizers have been proposed, with a majority employing zero-order approaches such as evolutionary or mutation-based algorithms \cite{marvel_automated_2009,kulk_evaluation_2011,wu_deep_2015,bruce_deep_2016,sohn_amortised_2016}, particle swarms \cite{krohling_gaussian_2004, kulk_evaluation_2011,bolet_exploration_2024} or Bayesian optimization \cite{calandra_bayesian_2016, mayr_skill-based_2022,berkenkamp_bayesian_2023,bolet_exploration_2024}.
First-order optimizers propose to leverage gradient information for fast, stable convergence. \Ac{dp} proposes to represent programs as \acp{dcg}, which permit gradient computation for program parameters via automatic differentiation \cite{baydin_automatic_2018,margossian_review_2019,blondel_elements_2024}. In robotics, \ac{dp} has been primarily used to optimize control parameters in conjunction with differentiable physics engines \cite{toussaint_differentiable_2018,degrave_differentiable_2019,hu_difftaichi_2019,qiao_scalable_2020,jatavallabhula_bayesian_2023}, or as differentiable policies in reinforcement learning \cite{wan_differentiable_2023,mora_pods_2021,kolaric_local_2020,wang_supervised_2024}. Our approach represents robot programs as \acp{dcg}, comprising differentiable planners and artificial neural networks, to afford first-order optimization of program parameters.

\subsubsection{Trajectory optimization}

Program parameter optimization has typically been considered separately from the optimization of motion trajectories. First-order methods such as CHOMP \cite{ratliff_chomp_2009} promise fast convergence due to the exploitation of gradient information \cite{osa_multimodal_2020,howell_trajectory_2022,howell_calipso_2023,xu_revisiting_2023,jin_safe_2024}. \ac{gpmp} \cite{mukadam_gaussian_2016,mukadam_continuous-time_2018,bhardwaj_differentiable_2020,cosier_unifying_2024} represents robot trajectories as a \ac{gp} and realizes optimization by iteratively minimizing an objective function comprising smoothness and collision constraints. We generalize \ac{dgpmp2} \cite{bhardwaj_differentiable_2020}, a first-order extension of \ac{gpmp}, to N-\acp{dof} serial kinematics, add additional constraints such as human demonstrations and integrate it as a gradient-preserving path planner inside a first-order program optimizer. To our knowledge, we contribute the first gradient-based framework for jointly optimizing robot program parameters and motion trajectories.

\begin{figure*}[h!]
	\centering
	\includegraphics[width=\textwidth]{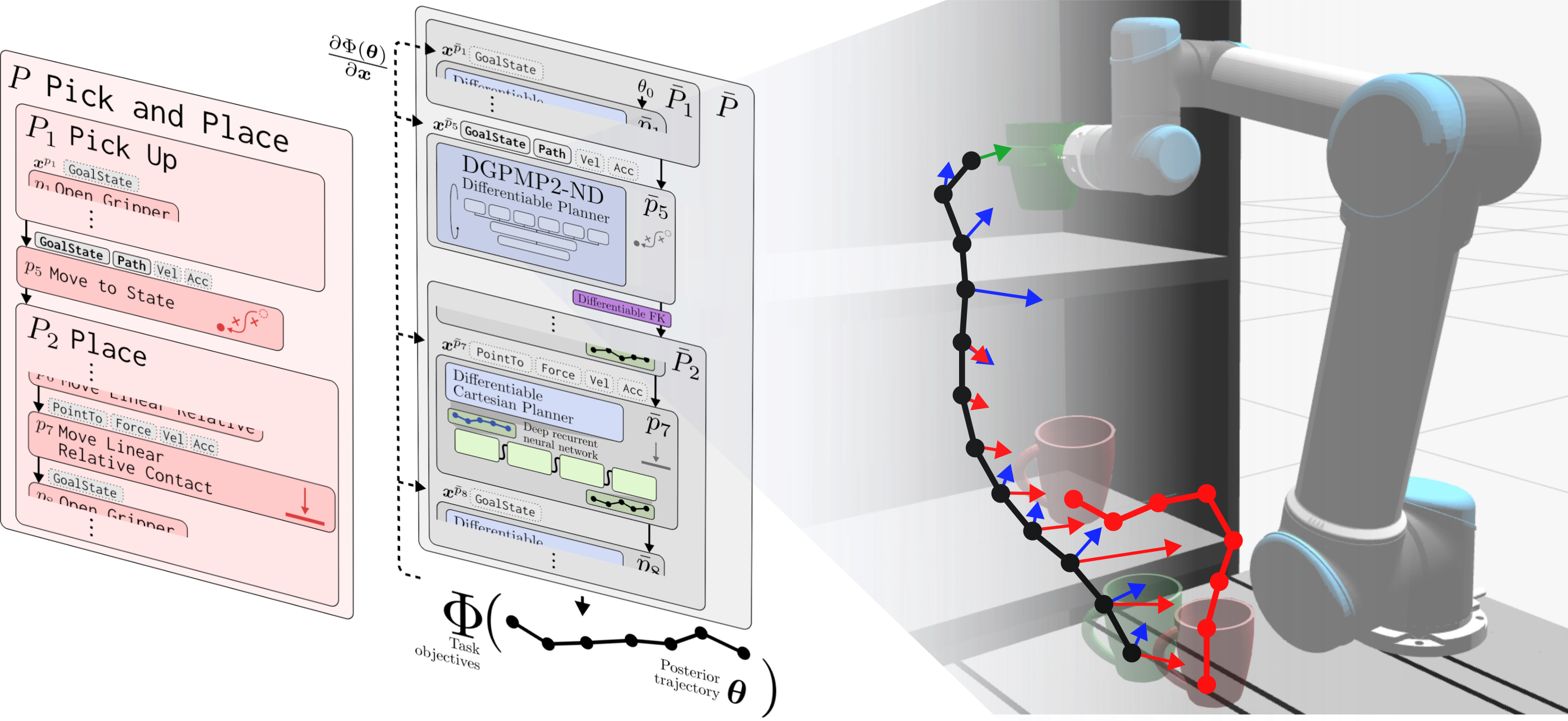}
	\caption{Left: \acf{spi-dp} optimizes skill-based robot programs (left, red) with respect to nearly arbitrary task objectives $\Phi$. Program parameters, such as the \texttt{Force} parameter of a placing action, can be optimized jointly with low-level motion trajectories to respect task-level objectives and motion-level constraints. By performing gradient-based optimization over a differentiable surrogate (``shadow'' $\bar{P}$, gray), the framework is applicable to near-arbitrary parameterized program representations, including most robot programming languages.
    Right: \ac{dgpmp2-nd} plans collision-free motions for N-\ac{dof} serial kinematics within \ac{spi-dp}'s optimization loop. Trajectories (black) are optimized with respect to collision (blue), goal (green), human demonstration (red) and other constraints.}
\end{figure*}

\section{Shadow Program Inversion: A Primer}

Our proposed framework is based on \acf{spi}, a model-based first-order optimizer for robot program parameters \cite{alt_robot_2021,alt_heuristic-free_2022,kienle_mutt_2024}, which is briefly outlined below. On the basis of \ac{spi} and a differentiable N-\ac{dof} motion planner (see Sec. \ref{sec:dgpmp}), we present a novel double-loop first-order optimizer capable of jointly optimizing program parameters and motion trajectories (see Sec. \ref{sec:joint-trajectory-and-parameter-optimization}).

\subsection{Differentiable Shadow Programs}
The core of our framework is the concept of a \textit{shadow program}: A differentiable ``twin'' of a robot program which serves as its surrogate for learning and optimization. The \textit{source program} and its shadow are representationally decoupled: The source program is written by the programmer and can be expressed in any parameterized representation, such as textual robot programming languages or task models \cite{schmidt-rohr_artiminds_2013,noauthor_simulator_2020,white_introducing_2023}, and is not required to be differentiable. The shadow program is a \ac{dcg}, which can be automatically constructed and partially learned \cite{alt_robot_2021}.

Given a source program $P(\bm{x},\theta_0)$ (such as an industrial robot program in a textual programming language), parameterized by program parameters $\bm{x}$ (such as waypoints, target poses, velocities or contact forces), we seek to optimize $\bm{x}$, given the initial robot state $\theta_0$. Mathematically, $P$ is a function $P: \mathbb{R}^N \times \Theta \rightarrow \Theta^T$, mapping a real-valued $N$-dimensional parameter vector $\bm{x} \in \mathbb{R}^N$ and joint angles $\theta$ in state space $\Theta$ to the robot trajectory $\bm{\theta} \in \Theta^T$, where $T$ is the number of timesteps. The state space is composed of robot joint configurations and end-effector wrenches: $\Theta = \mathcal{C} \times \mathbb{R}^6$.

The \textit{shadow program} $\bar{P}$ is a generative model of $P$, trained to approximate the real-world robot trajectory $\bm{\theta}$ for a given set of program parameters $\bm{x}$ and initial state $\theta_0$. The shadow program architecture is modular, to reflect the typically skill-based structure of most source programs, and permits $\bar{P}$ to be automatically constructed by traversing the source program and instantiating the corresponding shadow program skill by skill \cite{alt_robot_2021}. An exemplary shadow program composed of two skills is illustrated in Fig. \ref{fig:shadow-program}. For the purpose of this paper, it is important to note that shadow skills are \textit{generative models} of robot skills, predicting the expected real-world robot trajectory given the skill's parameters and the current robot state. A \textit{differentiable planner} bootstraps a naive \textit{prior trajectory}, which is then refined by a neural sequence-to-sequence model to reflect expected real-world deviations from the plan (see \cite{alt_robot_2021} and \cite{alt_heuristic-free_2022} for details).

A central property of $\bar{P}$ is that it is differentiable, allowing the computation of the gradient of a task-specific objective $\Phi$ over the trajectory with respect to the program's input parameters, and, as a consequence, the optimization of $\bm{x}$ using a gradient-based optimizer.

\subsection{Robot Program Parameter Optimization}

Differentiable shadow programs enable parameter optimization for near-arbitrary source programs in any parameterized representation. Consider a skill-based robot program (here in pseudocode) for an industrial peg-in-hole task: 
\begin{lstlisting}
MoveArm(approach_pose)
SpiralSearch(spiral_extents, contact_force)
Insert(depth, pushing_force)
\end{lstlisting}
\texttt{SpiralSearch} and \texttt{Insert} are skills from a skill library, and \texttt{approach\_pose} is a program parameter corresponding to the end-effector pose from which the robot starts searching for a hole in the workpiece. The cycle time of this program can be optimized by adapting \texttt{approach\_pose} to be, on average, directly above the hole. \ac{spi} solves such parameter optimization problems by first-order optimization over the shadow program, using $\bar{P}$ as a differentiable surrogate for $P$ \cite{alt_robot_2021}. The optimized parameters can be transferred back to the source program $P$, validated and adjusted by a human programmer, and executed on the robot.

\subsection{Joint Parameter and Trajectory Optimization}

For many tasks, program parameters and low-level motion trajectories must be jointly optimized. One example is the optimization of grasp poses to maximize grasp success: The approach and depart motions must also be optimized for collision-freeness, smoothness and other task constraints whenever the grasp pose is changed. One of the core contributions in this paper is a differentiable motion planner, integrated into the shadow program representation, which ensures that trajectories predicted in a forward pass comply with motion-level constraints such as collision-freeness, smoothness or proximity to a human demonstration. The differentiable motion planner is described in detail in Section \ref{sec:dgpmp}.

\section{DGPMP2-ND: Differentiable Motion Planning for N-DoF Manipulators}
\label{sec:dgpmp}

The gradient-based optimization of programs containing collision-free motion skills requires a differentiable planner. We propose \ac{dgpmp2-nd}, a differentiable collision-free motion planner for N-\ac{dof} manipulators, which generates trajectories that conform to motion constraints such as collision-freeness, smoothness, adherence to joint limits or precision at a target pose. To that end, we extend and modify \ac{dgpmp2} \cite{bhardwaj_differentiable_2020} by implementing differentiable collision checking for three-dimensional collision worlds and N-\ac{dof} serial kinematics, adding a joint limit constraint as well as a factor rewarding similarity to a human demonstration.

\subsection{Differentiable Gaussian Motion Planning}
\ac{dgpmp2} casts motion planning as inference on a factor graph \cite{bhardwaj_differentiable_2020} and minimizes a \textit{cost functional} $\mathcal{F}(\bm{\theta})$ over trajectory $\bm{\theta}$ via an iterative optimization procedure \cite{mukadam_continuous-time_2018}.

\begin{figure}
	\centering
	\includegraphics[width=\linewidth]{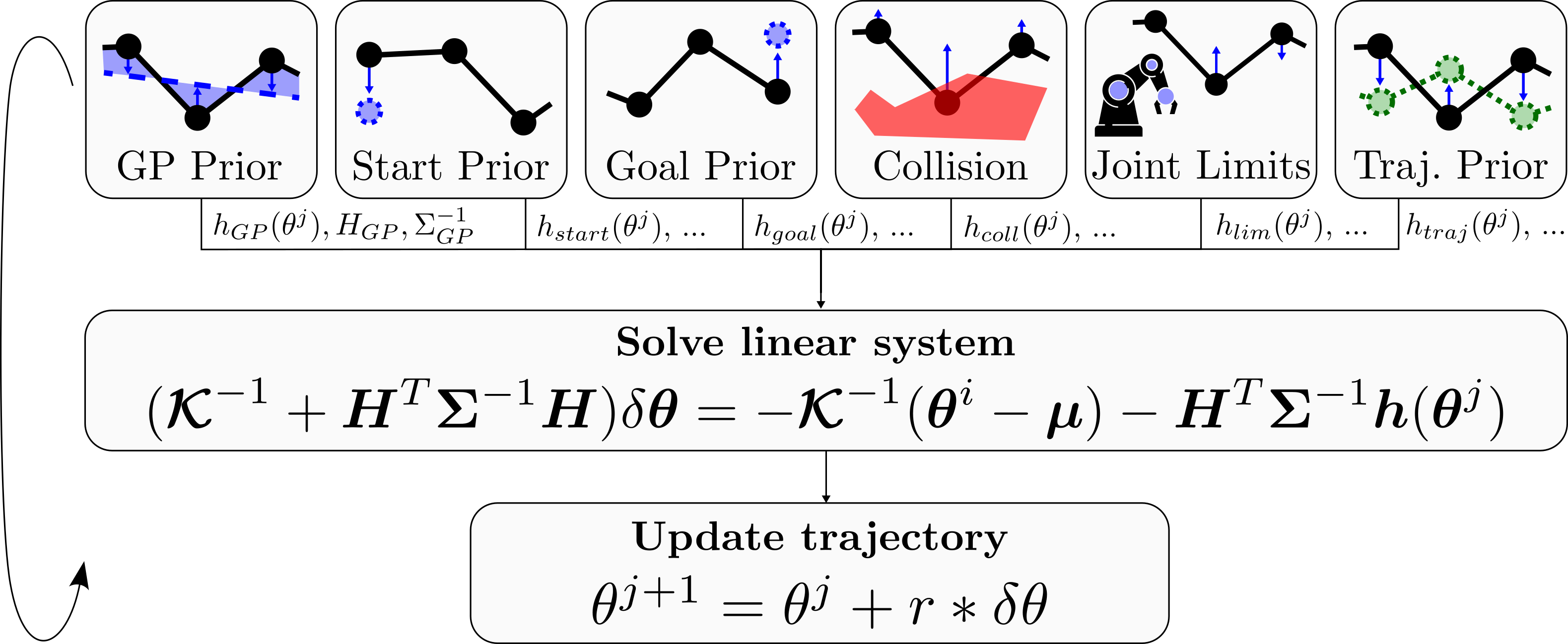}
	\caption{\acf{dgpmp2-nd} permits motion planning with respect to joint-space and Cartesian constraints and objectives. It realizes trajectory optimization by iterative solving of a linear system, permitting the backpropagation of gradients through the planner.}
	\label{fig:dgpmp2-nd}
\end{figure}

Figure \ref{fig:dgpmp2-nd} illustrates \ac{dgpmp2-nd}. While \ac{dgpmp2} plans in Cartesian space, \ac{dgpmp2-nd} plans joint-space trajectories. This permits to integrate joint limit constraints, while still supporting end-effector pose constraints by applying a differentiable forward kinematics on the joint-space trajectory. At each planner iteration $j, 1 \leq j \leq j_{max}$, a set of \textit{factors} is evaluated. Given the current joint trajectory $\bm{\theta}^j$, each factor computes an \textit{error} $\bm{h}(\bm{\theta})$, a \textit{Jacobian} $\hm{H}$ indicating the direction of steepest descent to minimize the error, and an \textit{inverse covariance} $\bm{\Sigma}^{-1}$ to weight the different factors. We propose six such factors:
\begin{enumerate}
	\item A \textit{\acf{gp} prior} factor, which penalizes points on the joint trajectory that deviate from the mean defined by a \ac{gp} prior (see \cite{mukadam_continuous-time_2018} for details). For each point on the trajectory, the Jacobian $\bm{H}_{GP}$ indicates the direction toward the GP mean.
	\item A \textit{start state prior}, which penalizes the deviation of the first point on $\bm{\theta}^j$ from a predefined start configuration. For the first point on $\bm{\theta}^j$, the Jacobian $\bm{H}_{start}$ indicates the direction toward the start configuration.
	\item A \textit{goal state prior}, which penalizes the deviation of the last point on $\bm{\theta}^j$ from a predefined goal configuration. For the last point on $\bm{\theta}^j$, the Jacobian $\bm{H}_{goal}$ indicates the direction toward the goal configuration.
	\item A \textit{collision factor} (see Sec. \ref{sec:collision-factor}).
	\item A \textit{joint limit factor} (see Sec. \ref{sec:joint-limit-factor}).
	\item A \textit{demonstration prior} (see Sec. \ref{sec:demonstration-prior}).
\end{enumerate}
The GP, start and goal state priors are identical to the original \ac{dgpmp2} formulation \cite{bhardwaj_differentiable_2020}, but are extended to the N-\ac{dof} case. We contribute novel differentiable collision and joint limit factors, as well as a differentiable Cartesian demonstration prior.

At each iteration, the linear system
\begin{equation}
	\begin{aligned}
		(\bm{\mathcal{K}}^{-1} + \bm{H}^{T} \bm{\Sigma}^{-1} \bm{H}) \delta\bm{\theta} = &- \bm{\mathcal{K}}^{-1}(\bm{\theta}^i - \bm{\mu})\\&- \bm{H}^{T} \bm{\Sigma}^{-1} \bm{h}(\bm{\theta}^{j})
	\end{aligned}
\end{equation}
is solved for $\delta\bm{\theta}$, where $\bm{H}$ is the combined Jacobian, $\bm{\Sigma^{-1}}$ is the combined inverse covariance, $\bm{\mathcal{K}}^{-1}$ is the inverse kernel matrix of the \ac{gp} and $\bm{h}(\bm{\theta}^{j})$ is the combined error function \cite{bhardwaj_differentiable_2020}. All matrices are combined by concatenating the matrices of the individual factors along the row axis. The trajectory is then incrementally updated: $\bm{\theta}^{j+1}=\bm{\theta}^j + r * \delta\bm{\theta}$, where $r$ is the update rate.

As \ac{dgpmp2-nd} is used as a differentiable planner inside another iterative optimization process, the total number of iterations required by \ac{dgpmp2-nd} until convergence must be kept as small as possible. To that end, $r$ is initalized at a high value ($r$=0.3), leaving it constant while the collision error $h_{coll} > 0$, and decay it by factor 0.1 at each iteration. Moreover, optimization is stopped before $j_{max}$ is reached when $h_{coll}$ has not decreased for 25 iterations or the total $h$ has decreased by less than 5\% for 25 iterations.

\subsection{Differentiable N-DoF Collision Factor}
\label{sec:collision-factor}
Bhardwaj et al. \cite{bhardwaj_differentiable_2020} propose a collision factor for 2D environments and a point robot. We extend their approach to 3D environments, joint-space trajectories, and N-DoF serial robot kinematics. Before planning, we precompute a 3D \ac{sdf} of the environment, where each voxel contains the signed distance from the voxel center to the next obstacle. For all states on joint trajectory $\bm{\theta}$, we compute the Cartesian poses and Jacobians of all links using differentiable forward kinematics \cite{sutanto_encoding_2020}. For each link and each time step, we identify the \ac{sdf} voxel that intersects with the collision mesh of the link and has the smallest distance to the collision environment. To compute a differentiable error, we then take for each identified voxel the weighted mean of the 26 surrounding voxels, resulting in a vector pointing away from the nearest collision. The resulting Jacobian equals the matrix multiplication of the Jacobian for each link and the Jacobian for the differentiable error.

\subsection{Joint Limit Factor}
\label{sec:joint-limit-factor}
To ensure that joint-limit constraints of the manipulator are met, we extend \ac{dgpmp2} by a joint limit factor. For each state on the joint trajectory $\bm{\theta}$, the \textit{joint limit error}
\begin{equation}
	h_{lim} = \begin{cases}
		\theta - \theta_{lim}    &\ \text{if\quad} \theta > \theta_{lim} \\
		-\theta_{lim} - \theta	 &\ \text{if\quad} \theta < -\theta_{lim} \\
		0						 &\ \text{otherwise}
	\end{cases}
\end{equation}
penalizes trajectory states which exceed or fall below the joint limits $\bm{\theta}_{lim}$. $\bm{H}_{lim}$ is the identity matrix for values outside the limits, zero otherwise.

\subsection{Demonstration Prior}
\label{sec:demonstration-prior}
For many planning problems, human demonstrations can be leveraged to guide the planner toward good solutions, speeding up convergence. We extend \ac{dgpmp2} by a prior factor which penalizes trajectories that deviate from a reference trajectory, such as a human demonstration. For every state on the joint trajectory $\bm{\theta}$, we compute the Cartesian end-effector pose $\bm{p}$ and Jacobian $\bm{H}_{traj}$. The demonstration prior error $\bm{h}_{traj}$ is the pointwise difference between $\bm{p}$ and the corresponding point on the reference trajectory.

Taken together, \ac{dgpmp2-nd} permits collision-free motion planning by iterative optimization, while respecting additional motion-level constraints such as joint limits or adherence to a reference trajectory. \ac{dgpmp2-nd} permits the differentiation of planned trajectories with respect to input parameters such as target poses, as well as the integration into second- or higher-order optimizers. In our implementation, gradients are computed via automatic differentiation \cite{paszke_automatic_2017}.

\section{Joint Trajectory and Parameter Optimization}
\label{sec:joint-trajectory-and-parameter-optimization}
\begin{figure}
	\centering
	\includegraphics[width=\linewidth]{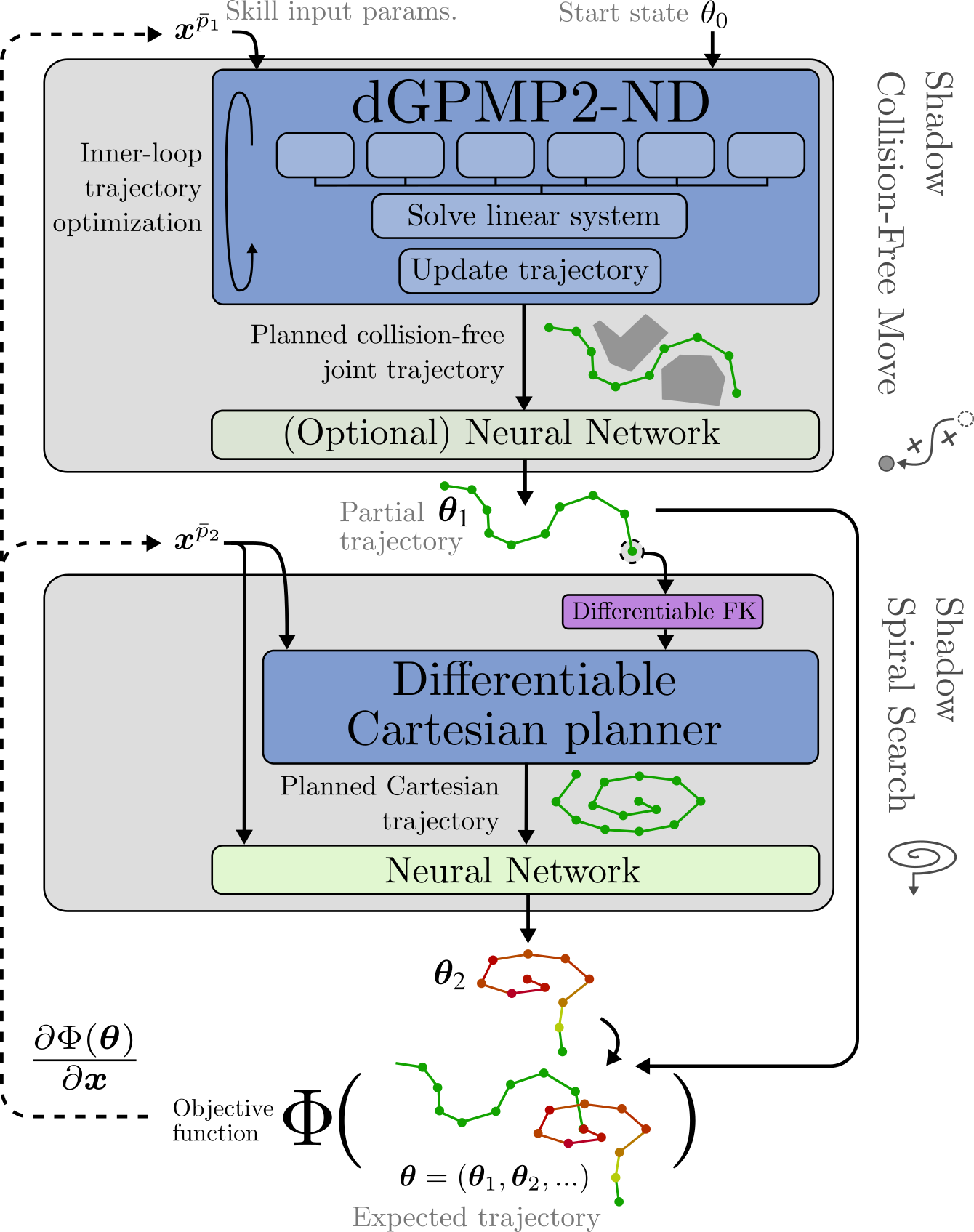}
	\caption{A differentiable shadow program for a search-based insertion task composed of two skills. By combining differentiable planners (blue) and trained neural networks (light green), program parameters $\bm{x}$ can be optimized with respect to task-level objectives $\Phi$ while respecting motion-level constraints such as collision-freeness. A forward pass (top to bottom) predicts the expected real-world trajectory given program parameters $\bm{x}$ and robot state $\theta$. The gradients of $\Phi$ are backpropagated and $\bm{x}$ is incrementally optimized.}
    \label{fig:shadow-program}
\end{figure}

For many real-world robot tasks, motion trajectories and program parameters cannot be optimized in isolation. Grasping is a canonical example: Grasping an object with a given grasp pose imposes constraints on the approach motion, while e.g. collision objects in the environment make some approach motions, and therefore grasp poses, impossible. Grasp poses and approach motions must be jointly optimized in order to achieve task-level objectives (a stable grasp) while obeying motion-level constraints (collision-free approach). With \ac{dgpmp2-nd}, gradient-based optimization over differentiable shadow programs permits the joint optimization of motion trajectories and program parameters. Fig. \ref{fig:shadow-program} shows the integration of \ac{dgpmp2-nd} as a differentiable collision-free motion planner into the shadow program architecture. Shadow programs are differentiable, predictive models of robot programs; with \ac{dgpmp2-nd}, collision-free planning is part of their forward pass.

\subsection{Shadow Program Inversion with Differentiable Planning}

With the integration of \ac{dgpmp2-nd} in the shadow program architecture, even complex multi-skill robot programs involving collision-free planning skills are represented as differentiable computation graphs. This enables the computation of $\frac{\partial \Phi(\bm{\theta})}{\partial \bm{x}}$, the gradient of some task-level objective function $\Phi$ of the predicted trajectory w.r.t. the program parameters $\bm{x}$, and the optimization of $\bm{x}$ by a first-order optimizer. We call this procedure \textit{\acf{spi-dp}}. For each iteration $i$, 
\begin{enumerate}
	\item a \textbf{forward pass} through the shadow program is performed, yielding a prediction of $\bm{\theta}$ given initial inputs $\bm{x}$ and start state $\theta_0$. This includes multiple iterations of \ac{dgpmp2-nd} as an inner-loop trajectory optimizer for each shadow skill involving $\mathcal{C}$-space planning;
	\item a \textbf{backward pass} is performed to compute $\frac{\partial \Phi(\bm{\theta})}{\partial \bm{x}}$ via automatic differentiation \cite{paszke_automatic_2017};
	\item the \textbf{input parameters are incrementally updated} via gradient descent to minimize $\Phi$. We use Adam \cite{kingma_adam_2015} with a relatively high learning rate, such as 0.001, for fast convergence. 
\end{enumerate}
Note that not all shadow skills require a computationally demanding planner such as \ac{dgpmp2-nd} for generating prior trajectories. For purely Cartesian skills such as spiral search \cite{alt_heuristic-free_2022} or linear motions, simple Cartesian planners suffice \cite{alt_robot_2021}. We find that the inner-loop \ac{dgpmp2-nd} converges in less than 20 iterations for most planning problems we encountered. In each outer-loop iteration $j$, planning results are cached and used as the initial trajectories for \ac{dgpmp2-nd} in the next iteration $j+1$, avoiding redundant planning. We find that one optimization iteration of a complex source program with 15 skills takes about 19 seconds on an Nvidia RTX 4090 \ac{gpu}.

\subsection{Task-Level Objective Functions}
As the learnable components of shadow skills are trained offline to accurately predict the expected trajectory, the task objective $\Phi$ does not need to be known at training time. Given trained shadow skills, program parameters can be optimized for near-arbitrary differentiable objective functions $\Phi$ over the expected trajectory $\bm{\theta}$. For industrial applications, the process metrics cycle time, path length and success probability are most salient:
\begin{align}
	\Phi_{cyc}(\bm{\theta}) &= \textstyle \sum_{i=1}^{|\bm{\theta}|} \textstyle \log(\bm{\theta}_{i,EOS})\\
	\Phi_{path}(\bm{\theta}) &= \textstyle \sum_{i=2}^{|\bm{\theta}|} \lVert {\bm{\theta}}_{i,pos} - \bm{\theta}_{i-1,pos}\rVert\\
	\Phi_{succ}(\bm{\theta}) &= \textstyle \sum_{i=1}^{|\bm{\theta}|} \textstyle \log(\bm{\theta}_{i,succ})
\end{align}

Both the cycle time $\Phi_{cyc}$ and the success probability $\Phi_{succ}$ are defined as the binary cross-entropy of the end-of-sequence and task success flags with a target label of $1$. For $\Phi_{succ}$, this pushes the success probability of every trajectory point to $1$ resulting in higher success probability of the execution. $\Phi_{cyc}$ pushes the end-of-sequence flag of every trajectory point toward $1$, such that the trajectory has fewer points, which is equivalent to a reduced cycle time. $\Phi_{path}$ calculates the overall path length of the trajectory, independent of the end-effector velocity.

With \ac{dgpmp2-nd}, \ac{spi-dp} optimizes program parameters subject to motion-level constraints such as collision-freeness, joint limits or proximity to a human demonstration (see \ref{sec:experiments-cupboard} and \ref{sec:experiments-engineblock} for details).

\section{Experiments}

\subsection{Household Pick-and-Place with Human Demonstration}
\label{sec:experiments-cupboard}

We evaluate our framework on a household table cleaning scenario, in which a robot is tasked to pick up a cup from a table and place it into a cupboard, while guaranteeing collision-freeness (see \ref{sec:cup-experiment-1}). The motions are conditioned on one single human demonstration. In a second set of experiments, we demonstrate the zero-shot transfer to a different object (a wine glass) and the simultaneous optimization of the target pose (see \ref{sec:cup-experiment-2}).
The experiments test three hypotheses:
\begin{enumerate}[label=H\arabic*]
\item \textbf{Motion-level optimization:} \ac{dgpmp2-nd} is capable of planning collision-free, smooth pick-and-place motions that adhere to a single human demonstration for variable target poses and different object geometries;
\item \textbf{Task-level optimization:} \ac{spi-dp} can optimize the entire robot program parameters with respect to task-level objectives, resulting in reduced overall cycle-time while respecting imposed contact-force limits;
\item \textbf{Joint optimization:} \ac{spi-dp} is capable of jointly optimizing robot programs with respect to motion-level (collision-freeness, human demonstration) and task-level (cycle time, contact force) constraints.
\end{enumerate}

The setup consists of a UR5 robotic arm with a flange-mounted ATI Gamma force-torque sensor and a SCHUNK pneumatic gripper. 10 demonstrations of a human transferring the cup from random pick-up poses to random target poses are collected with an Intel RealSense RGB-D camera. A human demonstration consists of the sampled 6D pose trajectory of the center of the cup.

\subsubsection{Cup Pick-and-Place}
\label{sec:cup-experiment-1}
\begin{figure}%
    \centering
    \includegraphics[width=\linewidth]{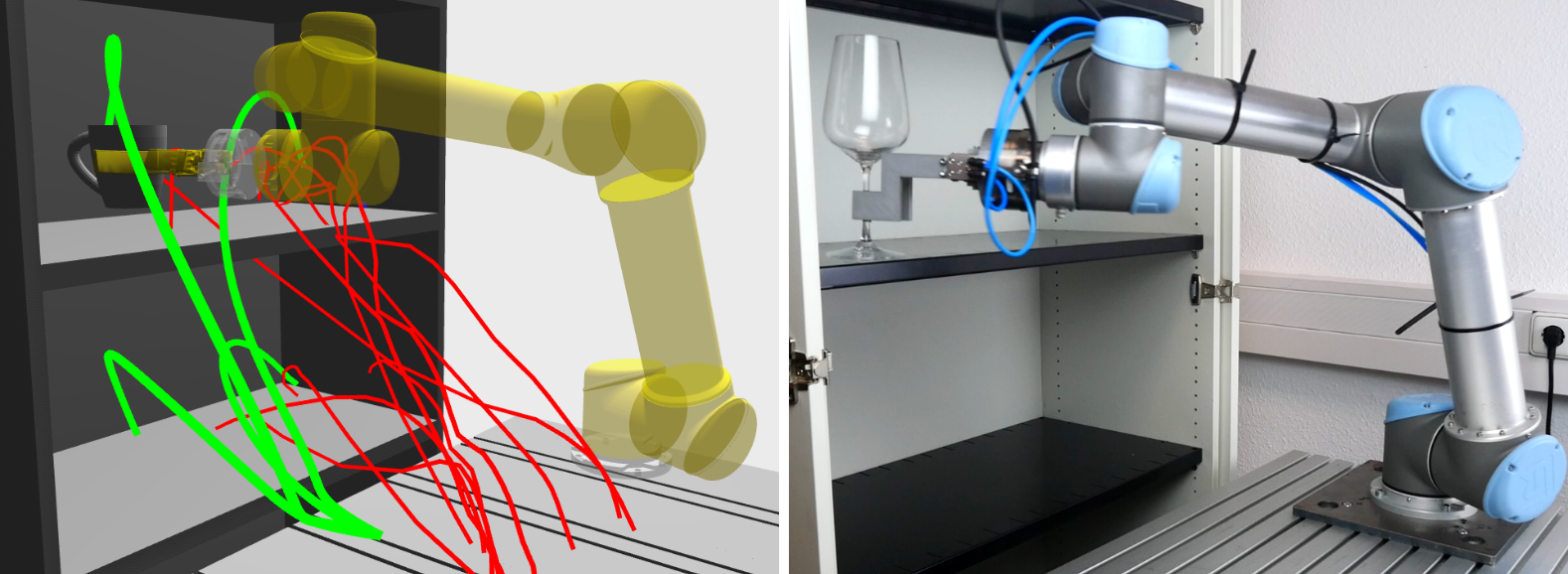}
    \caption{Left: Experiment \ref{sec:cup-experiment-1}: 3D rendering of the collision world, 4 exemplary optimization results (green) and 10 human demonstrations (red). Right: Experiment \ref{sec:cup-experiment-2}: Real-world execution of an optimized program.}%
    \label{fig:cupboard-results}%
\end{figure}

A robot program consisting of approach, grasp, transfer, place and depart skills is optimized to place the cup at one of four target poses on two different shelves inside the cupboard. In this experiment, the approach motion is optimized to respect the motion-level constraints illustrated in Fig. \ref{fig:dgpmp2-nd}, with a collision environment consisting of the robot, table, cup and cupboard, and a Cartesian trajectory prior given by a human demonstration. A total of 40 trials are performed, one for each combination of target pose and human demonstration. The results are shown in Fig. \ref{fig:cupboard-results} (left). All optimized motions were collision-free, even if the human demonstration contained a collision. The target pose was reached with a mean accuracy of 0.6 mm. The demonstration primarily acts as a regularizer that enforces implicit motion constraints, such as keeping the cup upright.

\subsubsection{Wine Glass Pick-and-Place with Target Pose Optimization}
\label{sec:cup-experiment-2}
The same robot program is optimized again, but the manipulated object is swapped for a wine glass and the gripper geometry is changed accordingly. In addition to the transfer motion, we also optimize the target pose (a parameter of the transfer skill) to minimize the cycle time of the overall program. The real-world experiment setup is shown in Fig. \ref{fig:cupboard-results} (right). Again, 40 trials are performed, one for each combination of initial target pose and human demonstration. All optimized motions were collision-free. \ac{spi-dp} optimized the target pose parameter of the transfer skill to be as close to the shelf as possible, yielding placing motions that were 2.8 cm shorter on average than before optimization. The placing motion has outsize impact on the cycle time due to its slow speed, which is required for force-controlled placing; overall, average cycle times were reduced by 40 \% while avoiding collisions.

\subsection{Engine Block Poka-Yoke Testing with Force Control}
\label{sec:experiments-engineblock}

\begin{figure}
    \centering
    \includegraphics[width=\linewidth]{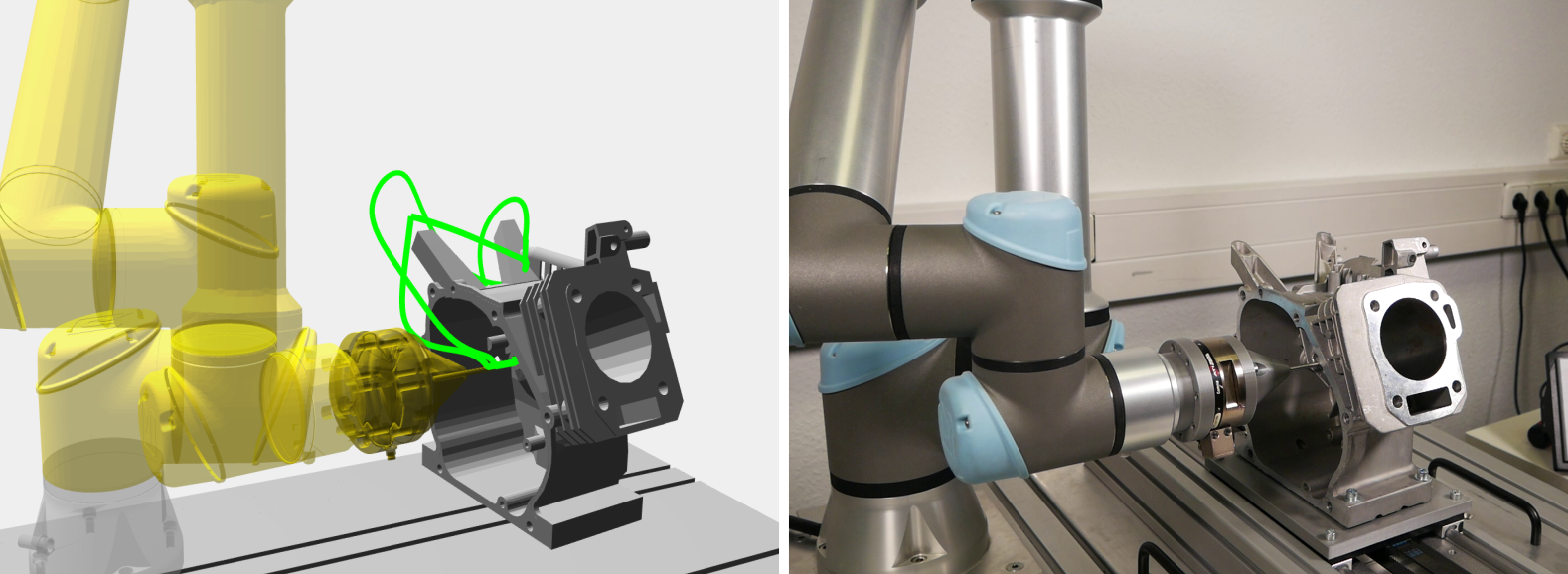}
    \caption{Experiment \ref{sec:experiments-engineblock}: 3D rendering of the collision world and collision free trajectory planned with \ac{spi-dp} (left) and real-world execution (right).}
    \label{fig:experiment-motorblock}
\end{figure}

This experiment tests the scalability of joint motion- and task-level optimization for complex industrial robot programs. The task consists of a poka-yoke quality assurance task, in which a UR5 robot arm approaches three holes on an engine block and executes a force-controlled spiral search motion to probe the hole (see Fig. \ref{fig:experiment-motorblock}. To simulate stochastic process noise, the engine block is moved on a linear axis by a random offset at every iteration. \ac{spi-dp} ensures collision-free motions from one hole to another and at the same time optimizes program parameters such as target poses, search patterns, velocities and contact forces with respect to the task-level objectives of cycle time minimization and maximization of the probability of task success (dropping into the hole before exausting the search pattern).

\begin{table}
    \begin{center}
        \caption{Experiment \ref{sec:experiments-engineblock}: Results}
        \label{tab:eval-engineblock}
        \begin{tabular}{lcccc}
            \hline
            &\multicolumn{2}{c}{Hole found} & \multicolumn{2}{c}{Duration (s)} \\
            &unoptimized &optimized & unoptimized & optimized\\
            \hline
            Hole 1	& 6\tiny{ / 20} & 11\tiny{ / 20} &2.29& 1.39\\
            Hole 2	& 6\tiny{ / 20}	& 10\tiny{ / 20} &1.95& 0.77\\
            Hole 3	& 7\tiny{ / 20}	& 20\tiny{ / 20} &1.97& 0.48\\
            \hline
        \end{tabular}
    \end{center}
\end{table}

20 trials are performed, for each of which the linear axis was moved randomly by up to 4 mm and the robot program was executed once with randomly initialized parameters, and once with optimized parameters. After optimization, the probability of finding each of the three holes is increased by 83\%, 67\% and 186\%, respectively (see Table \ref{tab:eval-engineblock}). In addition, optimization improved the search pattern, search dynamics and contact forces, reducing search duration by 62\%. Motions remained collision-free throughout. The noticeably better optimization results for hole 3 stem from its wider diameter and the greater surface area surrounding it, compared to the other holes.

\section{Conclusion and Outlook}
\label{sec:conclusion}

We introduce \acf{spi-dp}, a first-order robot program optimizer capable of jointly optimizing program parameters and motion trajectories. We present \ac{dgpmp2-nd}, a differentiable motion planner for N-\ac{dof} serial manipulators. \ac{spi-dp} leverages \ac{dgpmp2-nd} to optimize the parameters of robot programs with respect to task-level objectives, while simultaneously enforcing motion-level constraints. Experiments on two representative use cases from service and industrial robotics show that \ac{spi-dp} optimizes program parameters such as target poses or search regions while ensuring collision-freeness, smoothness and kinematic feasibility.
To our knowledge, \ac{spi-dp} is the first gradient-based optimizer capable of jointly optimizing program parameters and motion trajectories for arbitrary parameterized robot programs. Limitations include its relative sensitivity to hyperparameters, particularly the \ac{gp} and collision factor covariances. We suggest the future investigation of metaheuristics \cite{gendreau_handbook_2010,swan_metaheuristics_2022} and meta-optimization \cite{chen_online_2023,gautam_meta-learning_2023,feurer_initializing_2015} to steer the optimizer toward stable solutions or optimize hyperparameters for efficient convergence. Moreover, we seek to evaluate \ac{dgpmp2-nd} on large-scale planning benchmarks and investigate its integration as a differentiable planner for reinforcement learning \cite{wan_differentiable_2023,mora_pods_2021,kolaric_local_2020,wang_supervised_2024} and Task and Motion Planning \cite{envall_differentiable_2023,lee_stamp_2024,toussaint_differentiable_2018}.

\bibliographystyle{IEEEtran}  
\bibliography{bibliography}

\begin{thebibliography}{10}
\providecommand{\url}[1]{#1}
\csname url@rmstyle\endcsname
\providecommand{\newblock}{\relax}
\providecommand{\bibinfo}[2]{#2}
\providecommand\BIBentrySTDinterwordspacing{\spaceskip=0pt\relax}
\providecommand\BIBentryALTinterwordstretchfactor{4}
\providecommand\BIBentryALTinterwordspacing{\spaceskip=\fontdimen2\font plus
\BIBentryALTinterwordstretchfactor\fontdimen3\font minus \fontdimen4\font\relax}
\providecommand\BIBforeignlanguage[2]{{%
\expandafter\ifx\csname l@#1\endcsname\relax
\typeout{** WARNING: IEEEtran.bst: No hyphenation pattern has been}%
\typeout{** loaded for the language `#1'. Using the pattern for}%
\typeout{** the default language instead.}%
\else
\language=\csname l@#1\endcsname
\fi
#2}}

\bibitem{hoos_programming_2012}
H.~H. Hoos, ``Programming by optimization,'' \emph{Communications of the ACM}, vol.~55, no.~2, pp. 70--80, Feb. 2012.

\bibitem{ratliff_chomp_2009}
N.~Ratliff, M.~Zucker, J.~A. Bagnell, and S.~Srinivasa, ``{{CHOMP}}: {{Gradient}} optimization techniques for efficient motion planning,'' in \emph{2009 {{IEEE International Conference}} on {{Robotics}} and Automation ({{ICRA}})}.\hskip 1em plus 0.5em minus 0.4em\relax Kobe, Japan: IEEE, May 2009, pp. 489--494.

\bibitem{Kala11}
M.~Kalakrishnan, S.~Chitta, E.~Theodorou, P.~Pastor, and S.~Schaal, ``{{STOMP}}: {{Stochastic Trajectory Optimization}} for {{Motion Planning}},'' in \emph{2011 {{IEEE International Conference}} on {{Robotics}} and {{Automation}}}, May 2011, pp. 4569--4574.

\bibitem{osa_multimodal_2020}
T.~Osa, ``Multimodal trajectory optimization for motion planning,'' \emph{The International Journal of Robotics Research}, vol.~39, no.~8, pp. 983--1001, July 2020.

\bibitem{dastider_retro_2024}
A.~Dastider, H.~Fang, and M.~Lin, ``{{RETRO}}: {{Reactive Trajectory Optimization}} for {{Real-Time Robot Motion Planning}} in {{Dynamic Environments}},'' in \emph{2024 {{IEEE International Conference}} on {{Robotics}} and {{Automation}} ({{ICRA}})}, May 2024, pp. 8764--8770.

\bibitem{racca_interactive_2020}
M.~Racca, V.~Kyrki, and M.~Cakmak, ``Interactive {{Tuning}} of {{Robot Program Parameters}} via {{Expected Divergence Maximization}},'' in \emph{Proceedings of the 2020 {{ACM}}/{{IEEE International Conference}} on {{Human-Robot Interaction}}}, ser. {{HRI}} '20.\hskip 1em plus 0.5em minus 0.4em\relax New York, NY, USA: Association for Computing Machinery, Mar. 2020, pp. 629--638.

\bibitem{alt_robot_2021}
B.~Alt, D.~Katic, R.~J{\"a}kel, A.~K. Bozcuoglu, and M.~Beetz, ``Robot {{Program Parameter Inference}} via {{Differentiable Shadow Program Inversion}},'' in \emph{2021 {{IEEE International Conference}} on {{Robotics}} and {{Automation}} ({{ICRA}})}.\hskip 1em plus 0.5em minus 0.4em\relax Xi'an, China: IEEE, May 2021, pp. 4672--4678.

\bibitem{berkenkamp_bayesian_2023}
F.~Berkenkamp, A.~Krause, and A.~P. Schoellig, ``Bayesian optimization with safety constraints: Safe and automatic parameter tuning in robotics,'' \emph{Machine Learning}, vol. 112, no.~10, pp. 3713--3747, Oct. 2023.

\bibitem{kumar_practice_2024}
N.~Kumar, T.~Silver, W.~McClinton, L.~Zhao, S.~Proulx, T.~{Lozano-P{\'e}rez}, L.~Kaelbling, and J.~Barry, ``Practice {{Makes Perfect}}: {{Planning}} to {{Learning Skill Parameter Policies}},'' in \emph{Robotics: {{Science}} and {{Systems}} 2024}, Delft, Netherlands, July 2024.

\bibitem{marvel_automated_2009}
J.~A. Marvel, W.~S. Newman, D.~P. Gravel, G.~Zhang, {Jianjun Wang}, and T.~Fuhlbrigge, ``Automated learning for parameter optimization of robotic assembly tasks utilizing genetic algorithms,'' in \emph{2008 {{IEEE International Conference}} on {{Robotics}} and {{Biomimetics}}}, Feb. 2009, pp. 179--184.

\bibitem{kulk_evaluation_2011}
J.~Kulk and J.~S. Welsh, ``Evaluation of walk optimisation techniques for the {{NAO}} robot,'' in \emph{2011 11th {{IEEE-RAS International Conference}} on {{Humanoid Robots}}}, Oct. 2011, pp. 306--311.

\bibitem{wu_deep_2015}
F.~Wu, W.~Weimer, M.~Harman, Y.~Jia, and J.~Krinke, ``Deep {{Parameter Optimisation}},'' in \emph{Proceedings of the 2015 {{Annual Conference}} on {{Genetic}} and {{Evolutionary Computation}}}, ser. {{GECCO}} '15.\hskip 1em plus 0.5em minus 0.4em\relax New York, NY, USA: Association for Computing Machinery, July 2015, pp. 1375--1382.

\bibitem{bruce_deep_2016}
B.~R. Bruce, J.~M. Aitken, and J.~Petke, ``Deep {{Parameter Optimisation}} for {{Face Detection Using}} the {{Viola-Jones Algorithm}} in {{OpenCV}},'' in \emph{Search {{Based Software Engineering}}}, ser. Lecture {{Notes}} in {{Computer Science}}, F.~Sarro and K.~Deb, Eds.\hskip 1em plus 0.5em minus 0.4em\relax Cham: Springer International Publishing, 2016, pp. 238--243.

\bibitem{sohn_amortised_2016}
J.~Sohn, S.~Lee, and S.~Yoo, ``Amortised {{Deep Parameter Optimisation}} of {{GPGPU Work Group Size}} for {{OpenCV}},'' in \emph{Search {{Based Software Engineering}}}, ser. Lecture {{Notes}} in {{Computer Science}}, F.~Sarro and K.~Deb, Eds.\hskip 1em plus 0.5em minus 0.4em\relax Cham: Springer International Publishing, 2016, pp. 211--217.

\bibitem{krohling_gaussian_2004}
R.~Krohling, ``Gaussian swarm: A novel particle swarm optimization algorithm,'' in \emph{{{IEEE Conference}} on {{Cybernetics}} and {{Intelligent Systems}}, 2004.}, vol.~1, Dec. 2004, pp. 372--376 vol.1.

\bibitem{bolet_exploration_2024}
G.~Bolet, G.~Georgakoudis, K.~Parasyris, K.~W. Cameron, D.~Beckingsale, and T.~Gamblin, ``An {{Exploration}} of {{Global Optimization Strategies}} for {{Autotuning OpenMP-based Codes}},'' in \emph{2024 {{IEEE International Parallel}} and {{Distributed Processing Symposium Workshops}} ({{IPDPSW}})}, May 2024, pp. 741--750.

\bibitem{calandra_bayesian_2016}
R.~Calandra, A.~Seyfarth, J.~Peters, and M.~P. Deisenroth, ``Bayesian optimization for learning gaits under uncertainty,'' \emph{Annals of Mathematics and Artificial Intelligence}, vol.~76, no.~1, pp. 5--23, Feb. 2016.

\bibitem{mayr_skill-based_2022}
M.~Mayr, F.~Ahmad, K.~Chatzilygeroudis, L.~Nardi, and V.~Krueger, ``Skill-based {{Multi-objective Reinforcement Learning}} of {{Industrial Robot Tasks}} with {{Planning}} and {{Knowledge Integration}},'' \emph{arXiv:2203.10033 [cs]}, Mar. 2022.

\bibitem{baydin_automatic_2018}
A.~G. Baydin, B.~A. Pearlmutter, A.~A. Radul, and J.~M. Siskind, ``Automatic {{Differentiation}} in {{Machine Learning}}: {{A Survey}},'' \emph{Journal of Machine Learning Research}, vol.~18, no. 153, pp. 1--43, 2018.

\bibitem{margossian_review_2019}
C.~C. Margossian, ``A {{Review}} of automatic differentiation and its efficient implementation,'' \emph{WIREs Data Mining and Knowledge Discovery}, vol.~9, no.~4, July 2019.

\bibitem{blondel_elements_2024}
M.~Blondel and V.~Roulet, ``The {{Elements}} of {{Differentiable Programming}},'' Mar. 2024.

\bibitem{toussaint_differentiable_2018}
M.~Toussaint, K.~Allen, K.~Smith, and J.~Tenenbaum, ``Differentiable {{Physics}} and {{Stable Modes}} for {{Tool-Use}} and {{Manipulation Planning}},'' in \emph{Robotics: {{Science}} and {{Systems XIV}}}, vol.~14, June 2018.

\bibitem{degrave_differentiable_2019}
J.~Degrave, M.~Hermans, J.~Dambre, and F.~Wyffels, ``A {{Differentiable Physics Engine}} for {{Deep Learning}} in {{Robotics}},'' \emph{Frontiers in Neurorobotics}, vol.~13, 2019.

\bibitem{hu_difftaichi_2019}
Y.~Hu, L.~Anderson, T.-M. Li, Q.~Sun, N.~Carr, J.~{Ragan-Kelley}, and F.~Durand, ``{{DiffTaichi}}: {{Differentiable Programming}} for {{Physical Simulation}},'' in \emph{International {{Conference}} on {{Learning Representations}}}, Sept. 2019.

\bibitem{qiao_scalable_2020}
Y.-L. Qiao, J.~Liang, V.~Koltun, and M.~Lin, ``Scalable {{Differentiable Physics}} for {{Learning}} and {{Control}},'' in \emph{Proceedings of the 37th {{International Conference}} on {{Machine Learning}}}.\hskip 1em plus 0.5em minus 0.4em\relax PMLR, Nov. 2020, pp. 7847--7856.

\bibitem{jatavallabhula_bayesian_2023}
K.~M. Jatavallabhula, M.~Macklin, D.~Fox, A.~Garg, and F.~Ramos, ``Bayesian {{Object Models}} for {{Robotic Interaction}} with {{Differentiable Probabilistic Programming}},'' in \emph{Proceedings of {{The}} 6th {{Conference}} on {{Robot Learning}}}.\hskip 1em plus 0.5em minus 0.4em\relax PMLR, Mar. 2023, pp. 1563--1574.

\bibitem{wan_differentiable_2023}
W.~Wan, Y.~Wang, Z.~Erickson, and D.~Held, ``Differentiable {{Trajectory Optimization}} as a {{Policy Class}} for {{Reinforcement}} and {{Imitation Learning}},'' Oct. 2023.

\bibitem{mora_pods_2021}
M.~A.~Z. Mora, M.~Peychev, S.~Ha, M.~Vechev, and S.~Coros, ``{{PODS}}: {{Policy Optimization}} via {{Differentiable Simulation}},'' in \emph{Proceedings of the 38th {{International Conference}} on {{Machine Learning}}}.\hskip 1em plus 0.5em minus 0.4em\relax PMLR, July 2021, pp. 7805--7817.

\bibitem{kolaric_local_2020}
P.~Kolaric, D.~K. Jha, A.~U. Raghunathan, F.~L. Lewis, M.~Benosman, D.~Romeres, and D.~Nikovski, ``Local {{Policy Optimization}} for {{Trajectory-Centric Reinforcement Learning}},'' in \emph{2020 {{IEEE International Conference}} on {{Robotics}} and {{Automation}} ({{ICRA}})}, May 2020, pp. 5094--5100.

\bibitem{wang_supervised_2024}
L.~Wang, Y.~Zhang, D.~Zhu, S.~Coleman, and D.~Kerr, ``Supervised {{Meta-Reinforcement Learning With Trajectory Optimization}} for {{Manipulation Tasks}},'' \emph{IEEE Transactions on Cognitive and Developmental Systems}, vol.~16, no.~2, pp. 681--691, Apr. 2024.

\bibitem{howell_trajectory_2022}
T.~A. Howell, S.~Le~Cleac'h, S.~Singh, P.~Florence, Z.~Manchester, and V.~Sindhwani, ``Trajectory {{Optimization}} with {{Optimization-Based Dynamics}},'' \emph{IEEE Robotics and Automation Letters}, vol.~7, no.~3, pp. 6750--6757, July 2022.

\bibitem{howell_calipso_2023}
T.~A. Howell, K.~Tracy, S.~Le~Cleac'h, and Z.~Manchester, ``{{CALIPSO}}: {{A Differentiable Solver}} for~{{Trajectory Optimization}} with~{{Conic}} and~{{Complementarity Constraints}},'' in \emph{Robotics {{Research}}}, A.~Billard, T.~Asfour, and O.~Khatib, Eds.\hskip 1em plus 0.5em minus 0.4em\relax Cham: Springer Nature Switzerland, 2023, pp. 504--521.

\bibitem{xu_revisiting_2023}
M.~Xu, T.~L. Molloy, and S.~Gould, ``Revisiting {{Implicit Differentiation}} for {{Learning Problems}} in {{Optimal Control}},'' in \emph{Thirty-Seventh {{Conference}} on {{Neural Information Processing Systems}}}, Nov. 2023.

\bibitem{jin_safe_2024}
W.~Jin, S.~Mou, and G.~J. Pappas, ``Safe pontryagin differentiable programming,'' in \emph{Proceedings of the 35th {{International Conference}} on {{Neural Information Processing Systems}}}, ser. {{NIPS}} '21.\hskip 1em plus 0.5em minus 0.4em\relax Red Hook, NY, USA: Curran Associates Inc., June 2024, pp. 16\,034--16\,050.

\bibitem{mukadam_gaussian_2016}
M.~Mukadam, X.~Yan, and B.~Boots, ``Gaussian {{Process Motion}} planning,'' in \emph{2016 {{IEEE International Conference}} on {{Robotics}} and {{Automation}} ({{ICRA}})}, May 2016, pp. 9--15.

\bibitem{mukadam_continuous-time_2018}
M.~Mukadam, J.~Dong, X.~Yan, F.~Dellaert, and B.~Boots, ``Continuous-{{Time Gaussian Process Motion Planning}} via {{Probabilistic Inference}},'' \emph{The International Journal of Robotics Research}, vol.~37, no.~11, pp. 1319--1340, Sept. 2018.

\bibitem{bhardwaj_differentiable_2020}
M.~Bhardwaj, B.~Boots, and M.~Mukadam, ``Differentiable {{Gaussian Process Motion Planning}},'' in \emph{2020 {{IEEE International Conference}} on {{Robotics}} and {{Automation}} ({{ICRA}})}, May 2020, pp. 10\,598--10\,604.

\bibitem{cosier_unifying_2024}
L.~C. Cosier, R.~Iordan, S.~N.~T. Zwane, G.~Franzese, J.~T. Wilson, M.~Deisenroth, A.~Terenin, and Y.~Bekiroglu, ``A {{Unifying Variational Framework}} for {{Gaussian Process Motion Planning}},'' in \emph{Proceedings of {{The}} 27th {{International Conference}} on {{Artificial Intelligence}} and {{Statistics}}}.\hskip 1em plus 0.5em minus 0.4em\relax PMLR, Apr. 2024, pp. 1315--1323.

\bibitem{alt_heuristic-free_2022}
B.~Alt, D.~Katic, R.~J{\"a}kel, and M.~Beetz, ``Heuristic-{{Free Optimization}} of {{Force-Controlled Robot Search Strategies}} in {{Stochastic Environments}},'' in \emph{2022 {{IEEE}}/{{RSJ International Conference}} on {{Intelligent Robots}} and {{Systems}} ({{IROS}})}.\hskip 1em plus 0.5em minus 0.4em\relax Kyoto, Japan: IEEE, Oct. 2022, pp. 8887--8893.

\bibitem{kienle_mutt_2024}
C.~Kienle, B.~Alt, O.~Celik, P.~Becker, D.~Katic, R.~J{\"a}kel, and G.~Neumann, ``{{MuTT}}: {{A Multimodal Trajectory Transformer}} for {{Robot Skills}},'' in \emph{{{IEEE}}/{{RSJ International Conference}} on {{Intelligent Robots}} and {{Systems}} ({{IROS}})}.\hskip 1em plus 0.5em minus 0.4em\relax Abu Dhabi, United Arab Emirates: IEEE, Aug. 2024, pp. 9644--9651.

\bibitem{schmidt-rohr_artiminds_2013}
S.~R. {Schmidt-Rohr}, R.~J{\"a}kel, and G.~Dirschl, ``{{ArtiMinds Robot Programming Suite}},'' ArtiMinds Robotics GmbH, 2013.

\bibitem{noauthor_simulator_2020}
``Simulator for industrial robots and offline programming - {{RoboDK}},'' https://robodk.com/, May 2020.

\bibitem{white_introducing_2023}
W.~T. White, ``Introducing {{Intrinsic Flowstate}},'' May 2023.

\bibitem{sutanto_encoding_2020}
G.~Sutanto, A.~Wang, Y.~Lin, M.~Mukadam, G.~Sukhatme, A.~Rai, and F.~Meier, ``Encoding {{Physical Constraints}} in {{Differentiable Newton-Euler Algorithm}},'' in \emph{Proceedings of the 2nd {{Conference}} on {{Learning}} for {{Dynamics}} and {{Control}}}.\hskip 1em plus 0.5em minus 0.4em\relax PMLR, July 2020, pp. 804--813.

\bibitem{paszke_automatic_2017}
A.~Paszke, S.~Gross, S.~Chintala, G.~Chanan, E.~Yang, Z.~DeVito, Z.~Lin, A.~Desmaison, L.~Antiga, and A.~Lerer, ``Automatic differentiation in {{PyTorch}},'' in \emph{{{NIPS}} 2017 {{Workshop Autodiff}}}, Oct. 2017.

\bibitem{kingma_adam_2015}
D.~P. Kingma and J.~Ba, ``Adam: {{A Method}} for {{Stochastic Optimization}},'' in \emph{3rd {{International Conference}} for {{Learning Representations}}}.\hskip 1em plus 0.5em minus 0.4em\relax San Diego: arXiv, 2015.

\bibitem{gendreau_handbook_2010}
M.~Gendreau and J.-Y. Potvin, Eds., \emph{Handbook of {{Metaheuristics}}}, ser. International {{Series}} in {{Operations Research}} \& {{Management Science}}.\hskip 1em plus 0.5em minus 0.4em\relax Boston, MA: Springer US, 2010, vol. 146.

\bibitem{swan_metaheuristics_2022}
J.~Swan, S.~Adriaensen, A.~E.~I. Brownlee, K.~Hammond, C.~G. Johnson, A.~Kheiri, F.~Krawiec, J.~J. Merelo, L.~L. Minku, E.~{\"O}zcan, G.~L. Pappa, P.~{Garc{\'i}a-S{\'a}nchez}, K.~S{\"o}rensen, S.~Vo{\ss}, M.~Wagner, and D.~R. White, ``Metaheuristics ``{{In}} the {{Large}}'','' \emph{European Journal of Operational Research}, vol. 297, no.~2, pp. 393--406, Mar. 2022.

\bibitem{chen_online_2023}
X.~Chen and E.~Hazan, ``Online {{Control}} for {{Meta-optimization}},'' \emph{Advances in Neural Information Processing Systems}, vol.~36, pp. 36\,768--36\,780, Dec. 2023.

\bibitem{gautam_meta-learning_2023}
T.~Gautam, S.~Pfrommer, and S.~Sojoudi, ``Meta-{{Learning Parameterized First-Order Optimizers Using Differentiable Convex Optimization}},'' in \emph{2023 62nd {{IEEE Conference}} on {{Decision}} and {{Control}} ({{CDC}})}, Dec. 2023, pp. 2284--2291.

\bibitem{feurer_initializing_2015}
M.~Feurer, J.~Springenberg, and F.~Hutter, ``Initializing {{Bayesian Hyperparameter Optimization}} via {{Meta-Learning}},'' \emph{Proceedings of the AAAI Conference on Artificial Intelligence}, vol.~29, no.~1, Feb. 2015.

\bibitem{envall_differentiable_2023}
J.~Envall, R.~Poranne, and S.~Coros, ``Differentiable {{Task Assignment}} and {{Motion Planning}},'' in \emph{2023 {{IEEE}}/{{RSJ International Conference}} on {{Intelligent Robots}} and {{Systems}} ({{IROS}})}, Oct. 2023, pp. 2049--2056.

\bibitem{lee_stamp_2024}
Y.~Lee, P.~Huang, K.~M. Jatavallabhula, A.~Z. Li, F.~Damken, E.~Heiden, K.~Smith, D.~Nowrouzezahrai, F.~Ramos, and F.~Shkurti, ``{{STAMP}}: {{Differentiable Task}} and {{Motion Planning}} via {{Stein Variational Gradient Descent}},'' Jan. 2024.

\end{thebibliography}

\end{document}